\documentclass[11pt]{article}

\usepackage[final]{acl}

\usepackage{times}
\usepackage{latexsym}

\usepackage[T1]{fontenc}
\usepackage[utf8]{inputenc}
\usepackage{microtype}
\usepackage{inconsolata}
\usepackage{graphicx}
\usepackage[export]{adjustbox}
\usepackage{changepage}

\newif\ifdraft

\ifdraft
\newcommand{\KK}[1]{{\color{green}{\bf KK: #1}}}

\newcommand{\AF}[1]{{\color{blue}{\bf AF: #1}}}

\newcommand{\RZ}[1]{{\color{cyan}{\bf RZ: #1}}}

\newcommand{\rz}[1]{{\color{cyan} #1}}

\newcommand{\RI}[1]{{\color{brown}{\bf RI: #1}}}

\else
\newcommand{\KK}[1]{}

\newcommand{\AF}[1]{}

\newcommand{\RZ}[1]{}
\newcommand{\rz}[1]{#1}
\newcommand{\RI}[1]{}

\fi

\usepackage[most]{tcolorbox}

\usepackage{xcolor}
\definecolor{red}{HTML}{ea5545}
\definecolor{yellow}{HTML}{ebdc78}
\definecolor{green}{HTML}{87bc45}
\definecolor{blue}{HTML}{27aeef}
\definecolor{purple}{HTML}{b33dc6}

\usepackage{listings}

\definecolor{codegreen}{rgb}{0,0.6,0}
\definecolor{codegray}{rgb}{0.5,0.5,0.5}
\definecolor{codepurple}{rgb}{0.58,0,0.82}
\definecolor{backcolour}{rgb}{0.95,0.95,0.92}
\lstdefinestyle{mystyle}{
	backgroundcolor=\color{backcolour}, 
	commentstyle=\color{codegreen},
	keywordstyle=\color{magenta},
	numberstyle=\tiny\color{codegray},
	stringstyle=\color{codepurple},
	basicstyle=\ttfamily\footnotesize,
	breakatwhitespace=false,         
	breaklines=true,                 
	captionpos=b,                    
	keepspaces=true,                 
	numbers=none,                    
	numbersep=5pt,                  
	showspaces=false,                
	showstringspaces=false,
	showtabs=false,                  
	tabsize=2
}

\usepackage[frozencache,cachedir=.]{minted}
\tcbuselibrary{minted,breakable,xparse,skins}
\definecolor{bg}{gray}{0.95}

\DeclareTCBListing{mintedbox}{O{}m!O{}}{%
  breakable=true,
  listing engine=minted,
  listing only,
  minted language=#2,
  minted style=default,
  minted options={%
    gobble=0,
    breaklines=true,
    breakafter=,,
    fontsize=\small,
    numbersep=8pt,
    breaksymbol=\ ,
    #1},
  boxsep=0pt,
  left skip=0pt,
  right skip=0pt,
  left=10pt,
  right=10pt,
  top=3pt,
  bottom=3pt,
  arc=7pt,
  leftrule=0pt,
  rightrule=0pt,
  bottomrule=2pt,
  toprule=2pt,
  colback=bg,
  colframe=orange!70,
  enhanced,
  #3}

\newtcolorbox{benignbox1}{
  enhanced,
  colback=blue!10,
  colframe=blue!30!black,
  fonttitle=\bfseries,
  title=GPT generated Prompt Templates,
  sharp corners,
}

\newtcolorbox{benignbox2}{
  enhanced,
  colback=blue!10,
  colframe=blue!30!black,
  fonttitle=\bfseries,
  title=In-Context Learning Prompt Templates,
  sharp corners,
}

\newtcolorbox{benignbox3}{
  enhanced,
  colback=blue!10,
  colframe=blue!30!black,
  fonttitle=\bfseries,
  title= {\small \texttt{PII-Compass} demonstration},
  sharp corners,
}

\newtcolorbox{benignbox4}[1][]{ 
  enhanced,
  colback=blue!10,
  colframe=blue!30!black,
  fonttitle=\bfseries,
  sharp corners,
  title= {\small \texttt{#1}}, 
}

\usepackage[utf8]{inputenc} 
\usepackage[T1]{fontenc}    
\usepackage{hyperref}       
\usepackage{url}            
\usepackage{booktabs}       
\usepackage{amsfonts}       
\usepackage{nicefrac}       
\usepackage{microtype}      
\usepackage{lipsum}
\usepackage{graphicx}
\graphicspath{ {./images/} }

\usepackage{booktabs}
\usepackage{amsmath}
\usepackage{amssymb}
\usepackage{subcaption}
\usepackage{multirow}
\usepackage{multicol}
\usepackage{makecell}
\usepackage{bbm}
\usepackage[normalem]{ulem}

\usepackage{dialogue}

\makeatletter

\newcommand{\printfnsymbol}[1]{%
  \textsuperscript{\@fnsymbol{#1}}%
}

\usepackage[symbol]{footmisc}

\title{PrivacyScalpel: Enhancing LLM Privacy via Interpretable Feature Intervention with Sparse Autoencoders}

\author{
\textbf{Ahmed Frikha}\footnotemark[1] \quad
\textbf{Muhammad Reza Ar Razi}\footnotemark[1]  \\
\textbf{Krishna Kanth Nakka} \quad
\textbf{Ricardo Mendes \quad
Xue Jiang\quad Xuebing Zhou}\\
\,
Huawei Munich Research Center \\
\texttt{krishna.kanth.nakka@huawei.com}
}

\begin{document}
\maketitle

\footnotetext[1]{Equal Contribution}

\begin{abstract}

Large Language Models (LLMs) have demonstrated remarkable capabilities in natural language processing but also pose significant privacy risks by memorizing and leaking Personally Identifiable Information (PII). 
Existing mitigation strategies, such as differential privacy 
and neuron-level interventions,
often degrade model utility or fail to effectively prevent leakage. To address this challenge, we introduce \emph{PrivacyScalpel}, a novel privacy-preserving framework that leverages LLM interpretability techniques to identify and mitigate PII leakage while maintaining performance. PrivacyScalpel comprises three key steps: (1) \emph{Feature Probing}, which identifies layers in the model that encode PII-rich representations, (2) \emph{Sparse Autoencoding}, where a k-Sparse Autoencoder (k-SAE) disentangles and isolates privacy-sensitive features,
 and (3) \emph{Feature-Level Interventions}, which employ targeted ablation and vector steering to suppress PII leakage. 
 
Our empirical evaluation on {Gemma2-2b} and {Llama2-7b}, fine-tuned on the {Enron dataset}, shows that PrivacyScalpel significantly reduces email leakage from \textbf{5.15\%} to as low as \textbf{0.0\%}, while maintaining over \textbf{99.4\%} of the original model's utility. Notably, our method outperforms neuron-level interventions in privacy-utility trade-offs, demonstrating that acting on sparse, monosemantic features is more effective than manipulating polysemantic neurons. Beyond improving LLM privacy, our approach offers insights into the mechanisms underlying PII memorization, contributing to the broader field of model interpretability and secure AI deployment.
\end{abstract}


\section{Introduction}
\label{sec:introduction}

Large Language Models (LLMs) have achieved significant milestones in natural language processing (NLP), excelling in tasks such as text generation, question answering, and language translation \cite{brown2020language}. Despite their transformative capabilities, the training of LLMs on large-scale datasets introduces critical privacy concerns. Studies have shown that LLMs can memorize and output sensitive Personally Identifiable Information (PII), such as email addresses and phone numbers, when queried with adversarial prompts \cite{carlini2021extracting, lukas2023analyzing, nakka2024piicompass, nakka2024piiscope}. This PII leakage poses serious risks, particularly in applications like customer service chatbots, where user privacy is paramount. \cite{das2024securityprivacychallengeslarge}

Existing approaches to mitigating privacy leakage often rely on scrubbing the training data or leverage differential privacy techniques \cite{yu2021largescaleprivatelearning, lukas2023analyzing}. However, these methods come at the cost of model utility, limiting their applicability in performance-critical settings. Moreover, the underlying mechanisms through which LLMs memorize and leak sensitive information remain poorly understood, hindering the development of effective defenses. Concurrent works exploring neuron-level interventions to mitigate privacy leakage risks \cite{chen2024learnableprivacyneuronslocalization} have demonstrated potential but also suffer from significant performance degradation, further highlighting the need for solutions with better privacy-utility trade-offs.

To address these challenges, we propose \textbf{PrivacyScalpel}, a novel privacy-preserving framework that makes the following key contributions. First, PrivacyScalpel leverages recent interpretability techniques \cite{gao2024scalingevaluatingsparseautoencoders} to identify and isolate monosemantic features. These features represent distinct and interpretable concepts within the model's activations, enabling precise privacy-specific interventions. By acting on features directly responsible for PII leakage, our method reduces privacy risks without compromising downstream task performance. Second, we empirically demonstrate that acting on more disentangled and interpretable features is more effective in striking a good privacy-utility trade-off than manipulating polysemantic neuron-level activations \cite{bricken2023monosemanticity}. Third, our comprehensive evaluation on different models and datasets showcases the robustness and maturity of PrivacyScalpel for real-world applications. In particular, our approach fully mitigates email leakage while preserving a high performance on three benchmark Q\&A datasets. Finally, Beyond its empirical success, PrivacyScalpel offers insights into the internal mechanisms of PII memorization in LLMs, advancing both the understanding and mitigation of privacy risks in large-scale AI systems.


\section{Related Work}

Privacy concerns in machine learning, particularly in large language models (LLMs), have garnered significant attention due to their potential to memorize and inadvertently reveal sensitive information present in their training data \cite{carlini2021extracting}. This section discusses related work on privacy risks, interpretability in LLMs, and mitigation techniques, situating our contributions within this research landscape.

\subsection{Privacy Risks in LLMs}
Recent studies have highlighted the susceptibility of LLMs to privacy leakage through model memorization. For example, prior work showed that LLMs can memorize and output sensitive data, such as email addresses and social security numbers, when prompted with specific queries \cite{carlini2021extractingtrainingdatalarge}. This raises significant privacy concerns in applications involving user-generated or proprietary data, such as email processing or customer service chatbots. The development of benchmarks like TrustLLM \cite{huang2024trustllmtrustworthinesslargelanguage} and DecodingTrust \cite{wang2024decodingtrustcomprehensiveassessmenttrustworthiness} has further enabled systematic evaluation of privacy leakage in LLMs.

\subsection{Privacy-Preserving Methods in LLMs}

A variety of methods have been proposed to mitigate privacy risks in large language models (LLMs), focusing on different levels of intervention to balance privacy and utility. \citet{yu2021largescaleprivatelearning} introduced a low-rank reparameterization technique to address the scalability challenges of Differentially Private Stochastic Gradient Descent (DP-SGD) \AF{CITE DPSGD} \RZ{cited at beginning of sentence}. By decomposing weight matrices, this approach reduces memory overhead and noise intensity, enabling privacy-preserving training of large-scale models like BERT while achieving competitive utility scores. Similarly, \citet{chen2024learnableprivacyneuronslocalization} localized privacy-sensitive neurons using learnable binary masks, showing that PII is concentrated in specific neurons, particularly in Multi-Layer Perceptron (MLP) layers. Deactivating these neurons reduces privacy leakage but comes with a trade-off in model utility.

Other works have explored privacy-preserving mechanisms at different stages of the LLM pipeline. Tang et al. \cite{tong2023privacypreserving} proposed differentially private few-shot generation for in-context learning, creating synthetic demonstrations with formal privacy guarantees while retaining strong task performance. Wu et al. \cite{xinwei2023depn} presented DEPN, a framework for detecting and editing privacy neurons in pretrained language models, leveraging neuron-specific interventions to reduce leakage without significant utility loss. Majmudar et al. \cite{jimit2022differentially} extended privacy preservation to the decoding stage of LLMs, introducing a lightweight perturbation mechanism that applies differential privacy during text generation. 

Recent advances have also focused on structural properties of LLMs. Chen et al. \cite{tiejin2024privacypreserving} revealed that the flatness of the loss landscape in DP-trained models impacts the privacy-utility trade-off. They proposed a holistic framework leveraging weight flatness to improve generalization while maintaining differential privacy guarantees.
\RZ{fixed citation author's name}
Our work extends these efforts by focusing on feature-level interventions, such as sparse autoencoders and probing-based methods, to identify and mitigate privacy risks. Unlike neuron-specific approaches, our methodology leverages interpretability techniques to target privacy-relevant features directly, offering a robust trade-off between privacy preservation and model utility.

\subsubsection{Interpretability in LLMs Using Sparse Autoencoders}

Interpretability in large language models (LLMs) remains a fundamental challenge, as models often rely on polysemantic neurons that activate in multiple, semantically distinct contexts, making it difficult to understand their internal representations \cite{elhage2022superposition} \AF{CITE}. Sparse Autoencoders (SAEs) have emerged as a promising tool for disentangling these representations by learning sparse, monosemantic features that provide greater interpretability \cite{bricken2023monosemanticity}. \AF{CITE}

Cunningham et al. \cite{cunningham2023sparseautoencodershighlyinterpretable} demonstrated that SAEs can effectively resolve polysemanticity in LLM activations by learning sparse, human-interpretable features, which significantly improve the explainability of model behaviors. Building on this, Gao et al. \cite{gao2020pile800gbdatasetdiverse} explored the scalability of SAEs, introducing k-sparse autoencoders to directly control sparsity and improve the reconstruction-sparsity tradeoff. Their study also provided new evaluation metrics to assess feature quality, demonstrating that interpretability improves with autoencoder size.

Further extending this line of research, O'Neill and Bui \cite{oneill2024sparseautoencodersenablescalable} applied discrete sparse autoencoders to identify interpretable circuits in LLMs, showing that these methods allow efficient circuit discovery without requiring extensive ablations. Similarly, Rajamanoharan et al. \cite{rajamanoharan2024improvingdictionarylearninggated} introduced Gated Sparse Autoencoders, which mitigate the shrinkage effect of L1 penalties, leading to improved feature quality while maintaining interpretability. \RI{Describe what are interpretable circuits.}

Recent studies have also explored universality in feature representations across different LLMs. Lan et al. \cite{lan2024sparseautoencodersrevealuniversal} investigated how SAEs can reveal shared feature spaces across multiple LLM architectures, suggesting that interpretable features learned via SAEs are largely consistent across different models. Additionally, Marks et al. \cite{marks2024sparsefeaturecircuitsdiscovering} proposed sparse feature circuits, which use SAEs to discover causal subnetworks in LLMs, improving both interpretability and the ability to modify model behavior in a controlled manner.

Overall, these works highlight the potential of SAEs for making LLM activations more interpretable by transforming dense, polysemantic activations into sparse, monosemantic representations. Our work builds on these efforts by leveraging SAEs to identify privacy-relevant features and apply targeted interventions to mitigate privacy risks while maintaining model utility.
\RZ{add intepretability section to introduce SAE}

\section{Methodology}
\label{sec:methodology}

\begin{figure*}[h!]
    \centering
    \includegraphics[width=0.7\textwidth]{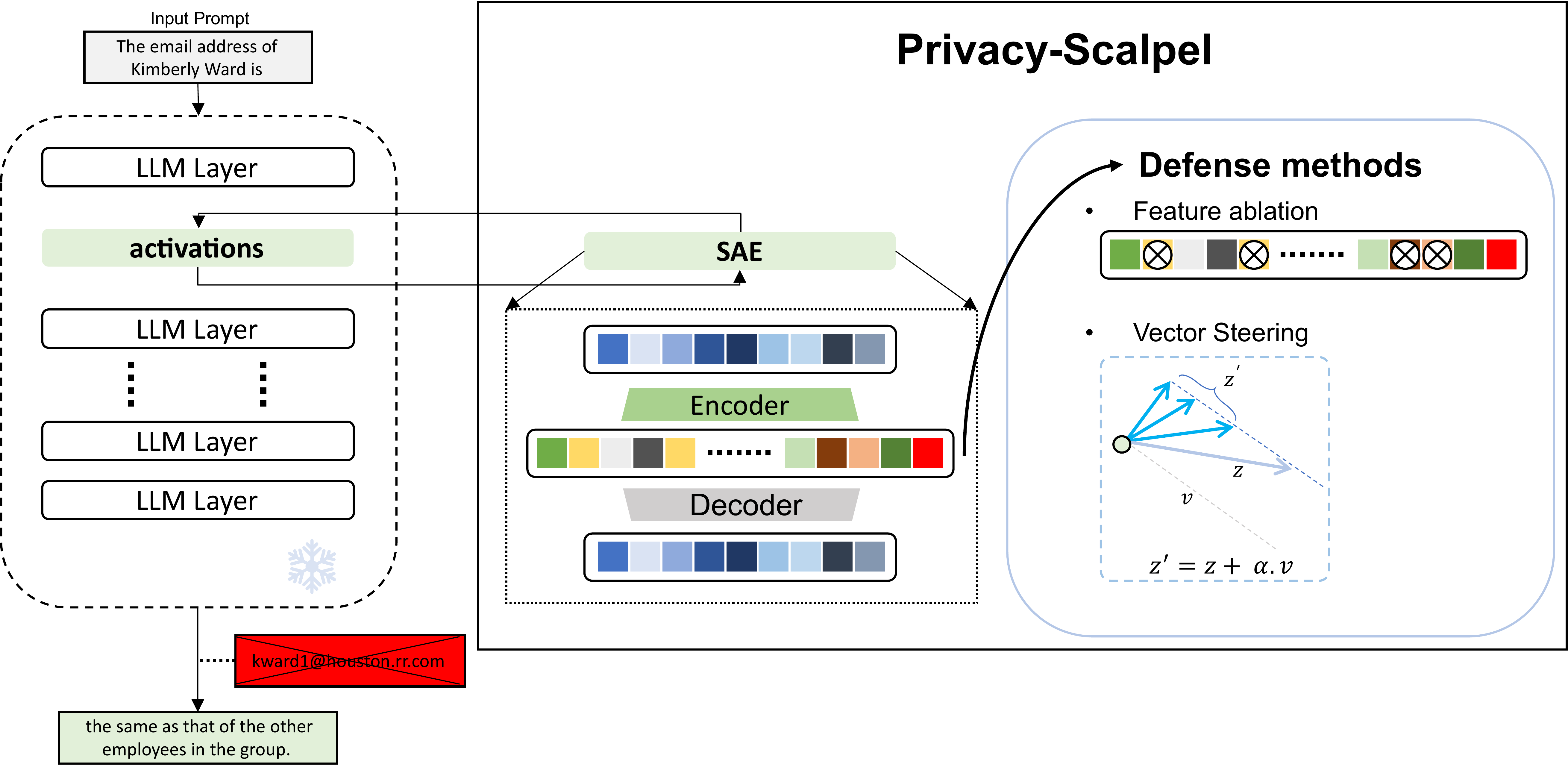} 
   \caption{{\bf Overview of our framework.} Given an input prompt, we extract the activations of input tokens at layer $l$ and perform intervention. We pass the activations through the SAE encoder and intervene on the SAE encoded high-dimensional features through Feature Ablation and Steering. We then decode the intervened features back to the original embedding space.
   }
    \label{fig:overview}
\end{figure*}

\noindent {\bf Problem Definition.}  
LLMs are trained on extensive datasets, which often contain sensitive data such as PII. This creates a critical privacy risk, as LLMs can memorize and inadvertently output sensitive information when queried with adversarial prompts \cite{carlini2021extracting, lukas2023analyzing}. To simulate a real-world scenario, we train an LLM on a dataset that includes sensitive data, reflecting practical use cases such as customer service chatbots and virtual assistants, where preserving user privacy is essential \cite{nakka2024piicompass, nakka2024piiscope}.

Formally, let $f_\theta$ denote an LLM parameterized by model parameters $\theta$, with an embedding dimension $d_{\text{emb}}$. Consider a set $S$ of data subjects, and for a specific subject $s_i \in S$, let $X_{\text{adv}}^i$ represent an adversarial prompt (e.g., "The email address of Karen Arnold is") targeting the leakage of a PII related to the data subject $s_i$. The adversarial prompt $X_{\text{adv}}^i$ consists of $T$ tokens $[x_1, x_2, \ldots, x_T]$. When prompted with $X_{\text{adv}}^i$, the LLM $f_\theta$ generates an output sequence $y = [x_{T+1}, \ldots, x_{T+N}]$, which may include memorized PII associated with $s_i$.  In the present work, we focus on email addresses.

The goal is to prevent the leakage of such sensitive information by intervening in the model activations, $A_k^l = [a_1^l, a_2^l, \ldots, a_k^l]$, during token generation at each timestep $k \in [T+1, N]$. Here, $A_k^l \in \mathbb{R}^d$ represents the embedding at layer $l$ for token $x_k$. The challenge is to design precise interventions that mitigate PII leakage while preserving the model’s performance on downstream tasks.

\noindent {\bf Overview.} PrivacyScalpel comprises three key steps. First, we probe across all layers to identify the optimal target layer $l$ that encodes the most PII-discriminative information, allowing us to pinpoint where interventions will be most effective (Sec~\ref{sec:probing}). Subsequently, we train a lightweight $k$-sparse autoencoder (k-SAE) \cite{makhzani2013k, gao2024scalingevaluatingsparseautoencoders} on the output residual stream of the target layer, mapping neuron embeddings to a more human-interpretable and high-dimensional disentangled feature space (Sec~\ref{sec:saetraining}). Lastly, we apply various intervention techniques on the neuron embeddings or within the feature space encoded by the k-SAE to effectively reduce PII leakage while maintaining model utility (Sec~\ref{sec:interventions}). \rz{An overview of this process is illustrated in Fig.~\ref{fig:overview}.}

\subsection{Probing for PII Features}\label{sec:probing}
In principle, causal interventions could be applied at all layers of the target LLM; however, this approach is computationally expensive. To efficiently mitigate privacy leakage, it is essential to identify the most suitable layer for intervention. To achieve this, we conduct a straightforward experiment to assess each layer's ability to distinguish between PII and non-PII data. 
\RZ{\rm{\sout{This evaluation helps us determine which layer will be most effective for interventions while remaining computationally feasible.}}}

Specifically, we train a classifier, referred to as a probe \cite{alain2018understandingintermediatelayersusing}, using the residual activations \( A^l \) from the transformer layers as input. The probe is designed to distinguish between sequences containing email addresses and those without. The model is formulated as 

\[
p_\theta(A^l) = \sigma(\langle \theta, A^l \rangle),
\]

where \( \theta \in \mathbb{R}^{d_{\text{emb}}} \) is the parameter vector, and \( A^l \) represents the residual activation at layer \( l \). Each transformer layer has its own probe, with \( A^l \) being the corresponding activation vector at that layer.

To train the probe, we use a labeled dataset \( D_{\text{prob}} = \{ X_{\text{PII}}, X_{\text{nonPII}} \} \), where \( X_{\text{PII}} \) contains sequences with email addresses, and \( X_{\text{nonPII}} \) contains sequences without any personally identifiable information (PII). The dataset is constructed from 1\% of the Pile dataset \cite{gao2020pile800gbdatasetdiverse}, taking only sequences that are less than 1024 tokens in length.  We then apply a regular expression to identify sequences containing email addresses and sample an equal number of sequences without email addresses to ensure balance in the dataset.  We defer more details about $D_{prob}$ to Appendix~\ref{sec:implementation}.

Each sequence is represented by a single aggregated activation vector, which is computed by averaging the residual activations for each sentence. This aggregated representation is used as the input to the classifier.

The classifier's performance is evaluated using validation accuracy for each layer, and the layer $l^{*}$ with the highest accuracy is selected as the target layer for further analysis. This process enables us to identify the transformer layer that most effectively captures PII-related information.

\subsection{Training the k-Sparse Autoencoder}\label{sec:saetraining} 
Once the target layer \( l \) has been selected based on probing results, we train a k-SAE on the activations \( A^l \) with a controlled level of sparsity. The k-SAE expands the input representation of dimension \( d_{\text{emb}} \) into latent features of dimension \( h \) using a TopK activation function \cite{gao2024scalingevaluatingsparseautoencoders} to ensure that only the top \( k \) largest activations are retained in the latent features.

Given an input activation \( a^l \), the encoder in k-SAE projects it into latent features \( z \in \mathbb{R}^h \) using:
\begin{equation}
z = \text{TopK}\left( W_{\text{enc}} \; (a^l - b_{\text{pre}}) \right)
\end{equation}\label{eq:encode}
where \( W_{\text{enc}} \in \mathbb{R}^{h \times d_{\text{emb}}} \) represents the encoder weight matrix, and \( b_{\text{pre}} \) is the "pre-encoder bias," which is subtracted from the input before encoding. The \( \text{TopK}(\cdot) \) function selects the largest \( k \) values from the resulting latent features, enforcing sparsity. 

The decoder then reconstructs the original activation \( a^l \) from the sparse latent features \( z \) using:
\begin{equation}
    \hat{a}^l = W_{\text{dec }}\;z + b_{\text{pre}}
\end{equation}
where \( W_{\text{dec}} \in \mathbb{R}^{d_{\text{emb}} \times h} \) is the decoder weight matrix. \RI{WD should have shape dem x H, if I am not mistaken. Confirm please.} \RZ{yes correct, forgot to update this}

The loss function \( \mathcal{L} \) for training the autoencoder is typically based on the mean squared error (MSE) between the original activation \( a^l \) and the reconstructed activation \( \hat{a}^l \):
\begin{equation}
    \mathcal{L} = \| a^l - \hat{a}^l \|^2
\end{equation}

\rz{To further improve the learned representations, an "auxiliary loss" \cite{gao2024scalingevaluatingsparseautoencoders} is introduced, which models the reconstruction error using latent features that have not been activated for a predefined number of tokens. This loss ensures that dead latents, which do not contribute to reconstruction, are also optimized. Specifically, given the reconstruction error of the main model \( e = a^l - \hat{a}^l \), the auxiliary loss is defined as:
\begin{equation}
    \mathcal{L}_{\text{aux}} = \| e - \hat{e} \|^2
\end{equation}
where \( \hat{e} = W_{D} z_{\text{aux}} \) is the reconstruction using the top \( k_{\text{aux}} \) inactive latents. The final loss function combines both components:
\begin{equation}
    \mathcal{L}_{\text{total}} = \mathcal{L} + \alpha \mathcal{L}_{\text{aux}}
\end{equation}
where \( \alpha \) is a small coefficient that controls the contribution of the auxiliary loss. This auxiliary loss helps mitigate feature collapse by ensuring that all latent dimensions contribute to the learned representations, improving the robustness of the sparse autoencoder.}

Overall, this training approach ensures that the reconstruction is accurate while enforcing sparsity in the latent features through the \( \text{TopK}(\cdot) \) activation function, which effectively limits the number of active latent units.

\subsection{Defense Method}\label{sec:interventions}
Building on this, PrivacyScalpel consists of two defense methods to mitigate privacy leakage while preserving model utility: feature ablation and feature steering. {\bf Feature ablation} removes the most privacy-sensitive latent features, while {\bf feature steering} modifies latent features to suppress PII-related information. By combining these approaches, PrivacyScalpel provides a flexible and effective privacy-preserving intervention.

\subsubsection{Feature Ablation}
\label{sec:feature_ablation}
To pinpoint the most active latent features associated with "PII features", particularly email addresses, we use a feature ablation technique, which we refer to as \emph{Ablation}. \rz{For this analysis, we utilize the dataset \(D_{\text{top-k}}\), which consists of 1538 sequences containing email addresses randomly sampled from the Enron dataset.} 

For each sequence containing an email address, we extract the corresponding  SAE latent features \( z \) using Eq.~\ref{eq:encode} from the model's activations at layer $l$, starting from the token where the email address first occurs and continuing until the end of the PII-containing segment. Here, \( z_i \) \RI{\( z_i \) or \( a_i \)?} represents the activation of the \( i \)-th latent feature in SAE space. These activations are aggregated across the sampled sequences and ranked by magnitude. The top \( k \) features with the highest magnitudes are then selected, based on the assumption that they are the most relevant for encoding PII.

\rz{After identifying the top $k$ features, \emph{Ablation} is performed by setting their activations to zero. The ablation is applied to the latent features of the last token at each timestep during generation, rather than across the entire input sequence. This approach is designed to minimize the propagated error introduced by the sparse autoencoder (SAE) reconstruction while still effectively suppressing privacy-sensitive features. By limiting the intervention to the last token, we ensure that the generated output is influenced by the privacy-preserving modification without unnecessarily disrupting the model’s overall performance on downstream tasks.}

\RI{Last token in the sequence just before the PII, right? Make this clear -- doesn't make sense that it is the last token of the sequence if the PII was already outputted
And if it is to the last token before the PII, is this sufficient? couldn't this lead to generating PII later? This needs to be clarified.Where is the proof/intuition/citation for being the most critical?}

\subsubsection{Feature Vector Steering}

To effectively alter latent features to achieve a desired outcome, we employ feature vector steering. In this approach, we use a steering vector, denoted as $v$, to modify the latent features through a linear transformation\cite{luo2024paceparsimoniousconceptengineering}. The adjusted SAE latent feature $z'$ is calculated as follows:

\begin{equation}
    z' = z + \alpha \cdot v
\end{equation}

where $z$ represents the original latent feature in SAE space, $v$ is the steering vector that captures the directional change in the feature space, and $\alpha$ is a scalar coefficient that controls the intensity of the adjustment.

To calculate the steering vector, we begin with dataset \rz{$D_{\text{prob}}$}, which contains text samples both with and without email addresses. \KK{How is D constructed, and what is its size? Is this $D_{\text{prob}}$ mentioned in Section 3.1} \RZ{yes, it's $D_{\text{prob}}$} For each sentence in the dataset, we collect the corresponding latent features and compute their average values. As a result, we represent each sentence with a single aggregated latent feature. 


We derive the steering vector $v$ using two methods: probing and the difference-in-means vectors.

\paragraph{Probing for Latent Features}
\label{sec:probing_features}

We refer to this method as \emph{Steering Probe}. In this approach, we train a probe using latent features \( z \) on the $D_{\text{prob}}$, where the parameters of the trained probe, denoted as \( \theta_z \), represent the direction of the PII feature in the latent space. After normalization, the vector \( \theta_z \) serves as the steering vector \( v \), which can influence the model’s behavior by shifting the latent feature \( z \) away from the direction corresponding to PII features.

\paragraph{Top-k Probing for Latent Features}

\rz{This method, which we refer to as \emph{Steering Top-k Probe}, builds upon the probing approach described in \emph{Steering Probe}, with a focus on the top $k$ selected latent features instead of all features. 
} 
We select top $k$ features using the feature ablation method discussed in Section~\ref{sec:feature_ablation}. After identifying the top $k$ features, a probe is trained on the $D_{\text{prob}}$ dataset, but only using the selected features. This allows for more efficient manipulation of the model, as the steering vector is applied only to the most active features, leaving the other latent dimensions unchanged.

\paragraph{Difference-in-Means Vectors}

We refer to this method as \emph{Steering Mean-Diff}. This approach calculates the steering vector using the difference-in-means technique. Inspired by previous work on steering models using a single direction \cite{eleuther_diff_in_means}, this method computes the difference between the mean activations of two sets of inputs: one containing PII and the other without PII. We use the same dataset as in probing. The steering vector is given by:
\begin{equation}
    v = \text{mean}(Z_{PII}) - \text{mean}(Z_{nonPII})
\end{equation}

where $Z_{PII}$ represents the latent features for inputs containing PII, and $Z_{nonPII}$ represents the latent features for inputs without PII. This vector can then be used to steer the latent feature representation towards the desired behavior.

The calculated steering vector is applied only to the latent features of the last token in the sequence. This is based on the observation that the final token is most critical in generating sensitive information, such as email addresses. Furthermore, to preserve the sparsity of the SAE latent features, the steering vector is added only to the active features—those that have nonzero activations. Non-active features, which are zero at the time, remain unchanged. This approach ensures that the sparse representation of the SAE latent space is maintained, reducing the risk of introducing noise into unrelated features.

By focusing the intervention on the last token and only modifying active features, this method minimizes disruption to the model's overall performance while effectively mitigating privacy leakage.

\section{Experiments}
\label{sec:experiments}

In this section, we evaluate the performance of PrivacyScalpel, our proposed privacy-enhancing toolbox, through a series of experiments designed to assess its effectiveness in reducing PII leakage while preserving model utility. The evaluation focuses on addressing three key research questions: (a) How effective are Sparse Autoencoders (SAEs) in mitigating PII leakage while preserving model utility? (b) How do feature-level interventions, such as SAE-based methods, compare to neuron-level interventions in terms of privacy preservation and utility? (c) How does the performance of privacy-preserving methods vary when trained on the full dataset (100\% of data) \KK{Which data?} \RZ{I explained it in \ref{sec:data-size-impact}.} compared to a significantly reduced dataset (1\% of data), particularly in terms of utility and effectiveness in mitigating PII leakage? These questions guide the design of our experiments and are referred to in the results discussion and experimental setup to provide clarity and structure.

\subsection{Experimental Setup}

\paragraph{Models.} 
To assess the performance of PrivacyScalpel, we conduct experiments using the \textbf{Gemma2-2b} \cite{gemmateam2024gemma2improvingopen} and \textbf{Llama2-7b} \cite{li2024llmpbeassessingdataprivacy} models, both fine-tuned on the Enron dataset, which contains real-world text data including PII.

\AF{we should mention llama here too, right? + cite it} \RI{Add details about the finetuning or cite the webpage where you got the model.}

\paragraph{Datasets.} 
We evaluate PrivacyScalpel using two types of datasets:
\begin{itemize}
    \item \textbf{Utility Evaluation Datasets:} These include:
    \begin{itemize}
        \item OpenBookQA \cite{OpenBookQA2018} for general knowledge reasoning.
        \item SciQ \cite{SciQ} for scientific question answering.
        \item PiQA \cite{Bisk2020} for physical reasoning tasks.
    \end{itemize}
    \item \textbf{Privacy Evaluation Dataset:} The \textbf{Adversarial Prompt Dataset (\(D_{\text{adv}}\))}, which contains prompts designed to elicit PII leakage. We discuss the detail of evaluation set in Appendix~\ref{sec:implementation}.
\end{itemize}
\RI{Food for thought: multi choice does not capture the loss in generative ability. We should consider some generative task to better measure utility, no?}

\paragraph{Evaluation Metrics.}
To evaluate PrivacyScalpel, we use the following metrics:
\begin{itemize}
    \item \textbf{Privacy Leakage:} This metric measures the proportion of prompts in \(D_{\text{adv}}\) where the model outputs the expected PII. Evaluation steps include:
    \begin{enumerate}
        \item Prompting the model with each sample from \(D_{\text{adv}}\).
        \item Comparing the model's output to the expected PII (e.g., email).
        \item Calculating the leakage rate as the percentage of prompts that result in correct PII extraction.
    \end{enumerate}
    \item \textbf{Utility Evaluation:} This metric assesses model performance on downstream tasks using PromptBench \cite{eval-harness}, which provides a unified framework for evaluating LLMs. The steps include:
    \begin{enumerate}
        \item Generating predictions for test samples from OpenBookQA, SciQ, and PiQA.
        \item Calculating the average accuracy across these datasets as a measure of model utility.
    \end{enumerate}
\end{itemize}

\subsection{Layer Selection for Intervention}
To determine the optimal layer for intervention, we probe the residual activations at each transformer layer using a classifier trained to distinguish between PII and non-PII data. As shown in Table~\ref{tab:probing_all_layers}, Layer 9 yields the highest validation accuracy and is selected as the target for subsequent analysis. Table~\ref{tab:sae_layer_comparison} further confirms these findings, as the application of Sparse Autoencoders (SAEs) at Layer 9 results in email leakage rates that closely match those of the original model without SAE. This indicates that Layer 9 best represents the PII features compared to other layers, as it retains the same level of leakage, demonstrating its alignment with the original model’s internal representations. By contrast, deeper layers such as Layer 20 show a significant reduction in leakage rates, suggesting that PII features become less prominent or undergo transformation as the representation progresses through the model. These results validate the selection of Layer 9 for capturing PII features effectively.

\begin{table*}[ht]
\centering
\small
\begin{tabular}{c|cc|cc|cc}
\toprule
\multirow{2}{*}{\textbf{Method}} & \multirow{2}{*}{\textbf{k}} & \multirow{2}{*}{\textbf{\( \alpha \)}} & \multicolumn{2}{c|}{\textbf{With SAE}} & \multicolumn{2}{c}{\textbf{Without SAE}} \\
\cline{4-7}
                                 &  & & \textbf{Avg. Utility} & \textbf{Email Leaks} & \textbf{Avg. Utility} & \textbf{Email Leaks} \\

\midrule
No defense & - & - & 58.52 & 5.15 & 58.77 & 5.15 \\
\hline
\multirow{4}{*} {Ablation}
                        & 100 & - & 58.54 & 3.72 & 58.1 & 4.58 \\
                        & 1000 & - & 58.25 & 0.03 & 58.07 & 2.83 \\
                        & 2000 & - & 58.05 & 0.01 & 55.4 & 0.01 \\
\hline
\multirow{4}{*} {Steering probe vector}
                        & - & -100.0 & 58.16 & 2.35 & 58.39 & 3.65 \\
                        & - & -200.0 & 57.94 & 0.04 & 57.26 & 0.28 \\
                        & - & -300.0 & 57.96 & 0.0 & 56.68 & 0.02 \\
\hline
\multirow{4}{*} {Steering topk-probe vector}
                        & - & -100.0 & 58.42 & 3.78 & 58.68 & 4.49 \\
                        & - & -200.0 & 58.51 & 0.22 & 58.39 & 1.14 \\
                        & - & -300.0 & 58.19 & 0.0 & 58.07 & 0.03 \\
\hline
\multirow{4}{*} {Steering mean-diff}
                        & - & -100.0 & 58.05 & 2.28 & 58.1 & 3.48 \\
                        & - & -200.0 & 56.6 & 0.0 & 57.25 & 0.0 \\
                        & - & -300.0 & 53.83 & 0.0 & 56.71 & 0.0 \\

\bottomrule
\end{tabular}
\caption{Comparison of defense performance for the Gemma2-2b model fine-tuned on the Enron dataset, evaluated under different configurations with and without K-SAE intervention in layer 9 (latent feature size = 65536). The table compares the effectiveness of various defense strategies, including TopK Ablation, vector steering, and difference-in-means (mean-diff) methods. 
}
\label{tab:results}
\end{table*}

\subsection{Effectiveness of Defense Methods}
This section evaluates the effectiveness of PrivacyScalpel's defense methods applied to the Gemma2-2b and Llama2-7b models fine-tuned on the Enron dataset, as shown in Tables~\ref{tab:results} and \ref{tab:results_llama2-7b}. These experiments assess the impact of various defense strategies, including \emph{Ablation}, \emph{Steering Probe}, \emph{Steering Top-k Probe}, \emph{Steering Mean-Diff}, both with and without Sparse Autoencoders (SAE). The results highlight consistent trends across both models, demonstrating the role of the SAE in enhancing privacy protection while maintaining utility.

For the Gemma2-2b model, \emph{Ablation} with SAE achieves significant leakage reduction, with leakage rates as low as 0.01\% (2,000 features ablated) while maintaining a utility score of 58.05\%. Without SAE, the same configuration results in a comparable leakage mitigation but also leads to a sharp utility drop to 55.40\%, underscoring the SAE's effectiveness in balancing privacy and utility. Similarly, \emph{Steering Probe} methods achieve zero leakage at high steering intensities (\(\alpha = -300.0\)) with SAE, while maintaining utility scores close to 57.96\%, outperforming configurations without SAE. \emph{Steering Top-k Probe} shows robust performance, achieving 0.0\% leakage with SAE (\(\alpha = -300.0\)) and utility scores of 58.19\%, highlighting its suitability for high-privacy requirements. \AF{I tried to add \% everywhere, check if I missed some to add them.}

For the Llama2-7b model, similar patterns emerge. \emph{Ablation} with SAE reduces leakage to 0.0\% (2,000 features ablated) while preserving a utility score of 64.60\%, compared to 63.48\% without SAE. Vector steering with SAE achieves zero leakage at \(\alpha = -30.0\), but at the cost of utility degradation, demonstrating a trade-off between privacy and utility. Across both models, SAE consistently enables more effective feature-level interventions, outperforming configurations without SAE in retaining utility while reducing leakage.

In summary, the results from both tables confirm the robustness of PrivacyScalpel's defense strategies. Incorporating SAE consistently improves the trade-off between privacy and utility, with TopK Ablation and steering probe vectors emerging as the most effective methods. These findings underscore the advantage of feature-level interventions in enhancing privacy preservation in language models.

\subsection{Impact of Data Size on Performance}
\label{sec:data-size-impact}

\rz{To investigate how data size influences the performance of privacy-preserving methods, we conduct experiments using the same datasets introduced in \emph{Ablation} method and \emph{Steering Probe} method. The first dataset, \( D_{\text{top-k}} \), is used for developing the \emph{Ablation} method by identifying the top-k latent features associated with PII. The second dataset, \( D_{\text{prob}} \), is used for developing the \emph{Steering Probe} method by determining the probing direction for intervention. To evaluate the performance of PrivacyScalpel under limited data availability, we apply a 1\% subsampling to these datasets while maintaining their original purpose. Table~\ref{tab:results_data_size} summarizes the average utility and email leakage rates across various defense methods under these conditions. This setup allows us to assess the robustness of PrivacyScalpel in identifying and mitigating PII leakage with reduced data.}\AF{it is not precise how they are derived. did you split randomly in two halves ? explain in a way that the reader can reproduce without having your code.}
\AF{use emph for method name as I did below. do it everywhere and also use the same names to intriduce these methods in their respective methodolgy subsections, e.g., we refer to this method as \emph{Ablation}.}

The results show that methods such as \emph{Ablation} and \emph{Steering Top-k Probe} are robust to data size reductions, maintaining low leakage rates with minimal utility loss. For instance, the \emph{Ablation} method achieves a leakage rate of 0.03\% with the full dataset and 0.05\% with the reduced dataset, while utility scores remain high. Similarly, \emph{Steering Top-k Probe} achieve zero leakage with both dataset sizes, though utility scores decrease slightly with reduced data.

In contrast, \emph{Steering Probe} and \emph{Steering Mean-Diff} methods exhibit higher sensitivity to reduced data size. \emph{Steering Probe} at \(\alpha = -250.0\) achieve zero leakage with the full dataset but show a leakage rate of 2.52\% with the reduced dataset, despite slightly improved utility. The \emph{Steering Mean-Diff} method consistently suppresses leakage but experiences significant utility degradation with reduced data.

Overall, these findings suggest that the choice of method should account for data size, with \emph{Ablation} and \emph{Steering Top-k Probe} emerging as the most robust options for maintaining privacy and utility across varying dataset sizes.

\section{Conclusion}
\label{sec:conclusion}

In this work, we introduced \emph{PrivacyScalpel}, a privacy-preserving framework that leverages LLM interpretability techniques to mitigate PII leakage while maintaining model utility. Unlike prior methods that rely on neuron-level interventions or differential privacy, our approach operates at the feature level, utilizing k-Sparse Autoencoders  to disentangle and suppress privacy-sensitive representations. Our results highlight that acting on sparse, monosemantic features is a more effective strategy for privacy preservation compared to manipulating polysemantic neurons \cite{bricken2023monosemanticity}. Additionally, our findings provide deeper insights into how LLMs encode and memorize sensitive information, contributing to the broader field of model interpretability and secure AI deployment.

\noindent{\bf Future Work.}  In future, we will explore extending PrivacyScalpel to mitigate other forms of sensitive information leakage beyond email addresses, such as financial records and personal identifiers. Additionally, integrating PrivacyScalpel with real-time inference settings could further enhance its applicability in privacy-sensitive domains such as healthcare and legal AI applications.  Overall, our work demonstrates that leveraging interpretability-driven interventions at the feature level provides a promising path forward for developing privacy-aware LLMs without significantly compromising their utility.

\bibliography{references}

\begin{thebibliography}{34}
\providecommand{\natexlab}[1]{#1}

\bibitem[{Alain and Bengio(2018)}]{alain2018understandingintermediatelayersusing}
Guillaume Alain and Yoshua Bengio. 2018.
\newblock \href {https://arxiv.org/abs/1610.01644} {Understanding intermediate layers using linear classifier probes}.
\newblock \emph{Preprint}, arXiv:1610.01644.

\bibitem[{Belrose(2023)}]{eleuther_diff_in_means}
Nora Belrose. 2023.
\newblock \href {https://blog.eleuther.ai/diff-in-means/} {Diff-in-means concept editing is worst-case optimal: Explaining a result by sam marks and max tegmark}.
\newblock Accessed: 2024-10-25.

\bibitem[{Bisk et~al.(2020)Bisk, Zellers, Bras, Gao, and Choi}]{Bisk2020}
Yonatan Bisk, Rowan Zellers, Ronan~Le Bras, Jianfeng Gao, and Yejin Choi. 2020.
\newblock Piqa: Reasoning about physical commonsense in natural language.
\newblock In \emph{Thirty-Fourth AAAI Conference on Artificial Intelligence}.

\bibitem[{Bricken et~al.(2023)Bricken, Templeton, Batson, Chen, Jermyn, Conerly, Turner, Anil, Denison, Askell, Lasenby, Wu, Kravec, Schiefer, Maxwell, Joseph, Hatfield-Dodds, Tamkin, Nguyen, McLean, Burke, Hume, Carter, Henighan, and Olah}]{bricken2023monosemanticity}
Trenton Bricken, Adly Templeton, Joshua Batson, Brian Chen, Adam Jermyn, Tom Conerly, Nick Turner, Cem Anil, Carson Denison, Amanda Askell, Robert Lasenby, Yifan Wu, Shauna Kravec, Nicholas Schiefer, Tim Maxwell, Nicholas Joseph, Zac Hatfield-Dodds, Alex Tamkin, Karina Nguyen, Brayden McLean, Josiah~E Burke, Tristan Hume, Shan Carter, Tom Henighan, and Christopher Olah. 2023.
\newblock Towards monosemanticity: Decomposing language models with dictionary learning.
\newblock \emph{Transformer Circuits Thread}.
\newblock Https://transformer-circuits.pub/2023/monosemantic-features/index.html.

\bibitem[{Brown(2020)}]{brown2020language}
Tom~B Brown. 2020.
\newblock Language models are few-shot learners.
\newblock \emph{arXiv preprint arXiv:2005.14165}.

\bibitem[{Carlini et~al.(2021{\natexlab{a}})Carlini, Tramer, Wallace, Jagielski, Herbert-Voss, Lee, Roberts, Brown, Song, Erlingsson, Oprea, and Raffel}]{carlini2021extractingtrainingdatalarge}
Nicholas Carlini, Florian Tramer, Eric Wallace, Matthew Jagielski, Ariel Herbert-Voss, Katherine Lee, Adam Roberts, Tom Brown, Dawn Song, Ulfar Erlingsson, Alina Oprea, and Colin Raffel. 2021{\natexlab{a}}.
\newblock \href {https://arxiv.org/abs/2012.07805} {Extracting training data from large language models}.
\newblock \emph{Preprint}, arXiv:2012.07805.

\bibitem[{Carlini et~al.(2021{\natexlab{b}})Carlini, Tramer, Wallace, Jagielski, Herbert-Voss, Lee, Roberts, Brown, Song, Erlingsson et~al.}]{carlini2021extracting}
Nicholas Carlini, Florian Tramer, Eric Wallace, Matthew Jagielski, Ariel Herbert-Voss, Katherine Lee, Adam Roberts, Tom Brown, Dawn Song, Ulfar Erlingsson, et~al. 2021{\natexlab{b}}.
\newblock Extracting training data from large language models.
\newblock In \emph{30th USENIX Security Symposium (USENIX Security 21)}, pages 2633--2650.

\bibitem[{Chen et~al.(2024{\natexlab{a}})Chen, Hu, Feng, and Liu}]{chen2024learnableprivacyneuronslocalization}
Ruizhe Chen, Tianxiang Hu, Yang Feng, and Zuozhu Liu. 2024{\natexlab{a}}.
\newblock \href {https://arxiv.org/abs/2405.10989} {Learnable privacy neurons localization in language models}.
\newblock \emph{Preprint}, arXiv:2405.10989.

\bibitem[{Chen et~al.(2024{\natexlab{b}})Chen, Da, Zhou, Li, Zhou, Chen, and Wei}]{tiejin2024privacypreserving}
Tiejin Chen, Longchao Da, Huixue Zhou, Pingzhi Li, Kaixiong Zhou, Tianlong Chen, and Hua Wei. 2024{\natexlab{b}}.
\newblock \href {https://arxiv.org/abs/2403.04124} {Privacy-preserving fine-tuning of large language models through flatness}.
\newblock \emph{Preprint}, arXiv:2403.04124.

\bibitem[{Cunningham et~al.(2023)Cunningham, Ewart, Riggs, Huben, and Sharkey}]{cunningham2023sparseautoencodershighlyinterpretable}
Hoagy Cunningham, Aidan Ewart, Logan Riggs, Robert Huben, and Lee Sharkey. 2023.
\newblock \href {https://arxiv.org/abs/2309.08600} {Sparse autoencoders find highly interpretable features in language models}.
\newblock \emph{Preprint}, arXiv:2309.08600.

\bibitem[{Das et~al.(2024)Das, Amini, and Wu}]{das2024securityprivacychallengeslarge}
Badhan~Chandra Das, M.~Hadi Amini, and Yanzhao Wu. 2024.
\newblock \href {https://arxiv.org/abs/2402.00888} {Security and privacy challenges of large language models: A survey}.
\newblock \emph{Preprint}, arXiv:2402.00888.

\bibitem[{Elhage et~al.(2022)Elhage, Hume, Olsson, Schiefer, Henighan, Kravec, Hatfield-Dodds, Lasenby, Drain, Chen, Grosse, McCandlish, Kaplan, Amodei, Wattenberg, and Olah}]{elhage2022superposition}
Nelson Elhage, Tristan Hume, Catherine Olsson, Nicholas Schiefer, Tom Henighan, Shauna Kravec, Zac Hatfield-Dodds, Robert Lasenby, Dawn Drain, Carol Chen, Roger Grosse, Sam McCandlish, Jared Kaplan, Dario Amodei, Martin Wattenberg, and Christopher Olah. 2022.
\newblock Toy models of superposition.
\newblock \emph{Transformer Circuits Thread}.
\newblock Https://transformer-circuits.pub/2022/toy\_model/index.html.

\bibitem[{Gao et~al.(2020)Gao, Biderman, Black, Golding, Hoppe, Foster, Phang, He, Thite, Nabeshima, Presser, and Leahy}]{gao2020pile800gbdatasetdiverse}
Leo Gao, Stella Biderman, Sid Black, Laurence Golding, Travis Hoppe, Charles Foster, Jason Phang, Horace He, Anish Thite, Noa Nabeshima, Shawn Presser, and Connor Leahy. 2020.
\newblock \href {https://arxiv.org/abs/2101.00027} {The pile: An 800gb dataset of diverse text for language modeling}.
\newblock \emph{Preprint}, arXiv:2101.00027.

\bibitem[{Gao et~al.(2024{\natexlab{a}})Gao, la~Tour, Tillman, Goh, Troll, Radford, Sutskever, Leike, and Wu}]{gao2024scalingevaluatingsparseautoencoders}
Leo Gao, Tom~Dupré la~Tour, Henk Tillman, Gabriel Goh, Rajan Troll, Alec Radford, Ilya Sutskever, Jan Leike, and Jeffrey Wu. 2024{\natexlab{a}}.
\newblock \href {https://arxiv.org/abs/2406.04093} {Scaling and evaluating sparse autoencoders}.
\newblock \emph{Preprint}, arXiv:2406.04093.

\bibitem[{Gao et~al.(2024{\natexlab{b}})Gao, Tow, Abbasi, Biderman, Black, DiPofi, Foster, Golding, Hsu, Le~Noac'h, Li, McDonell, Muennighoff, Ociepa, Phang, Reynolds, Schoelkopf, Skowron, Sutawika, Tang, Thite, Wang, Wang, and Zou}]{eval-harness}
Leo Gao, Jonathan Tow, Baber Abbasi, Stella Biderman, Sid Black, Anthony DiPofi, Charles Foster, Laurence Golding, Jeffrey Hsu, Alain Le~Noac'h, Haonan Li, Kyle McDonell, Niklas Muennighoff, Chris Ociepa, Jason Phang, Laria Reynolds, Hailey Schoelkopf, Aviya Skowron, Lintang Sutawika, Eric Tang, Anish Thite, Ben Wang, Kevin Wang, and Andy Zou. 2024{\natexlab{b}}.
\newblock \href {https://doi.org/10.5281/zenodo.12608602} {A framework for few-shot language model evaluation}.

\bibitem[{Huang et~al.(2024)Huang, Sun, Wang, Wu, Zhang, Li, Gao, Huang, Lyu, Zhang, Li, Liu, Liu, Wang, Zhang, Vidgen, Kailkhura, Xiong, Xiao, Li, Xing, Huang, Liu, Ji, Wang, Zhang, Yao, Kellis, Zitnik, Jiang, Bansal, Zou, Pei, Liu, Gao, Han, Zhao, Tang, Wang, Vanschoren, Mitchell, Shu, Xu, Chang, He, Huang, Backes, Gong, Yu, Chen, Gu, Xu, Ying, Ji, Jana, Chen, Liu, Zhou, Wang, Li, Zhang, Wang, Xie, Chen, Wang, Liu, Ye, Cao, Chen, and Zhao}]{huang2024trustllmtrustworthinesslargelanguage}
Yue Huang, Lichao Sun, Haoran Wang, Siyuan Wu, Qihui Zhang, Yuan Li, Chujie Gao, Yixin Huang, Wenhan Lyu, Yixuan Zhang, Xiner Li, Zhengliang Liu, Yixin Liu, Yijue Wang, Zhikun Zhang, Bertie Vidgen, Bhavya Kailkhura, Caiming Xiong, Chaowei Xiao, Chunyuan Li, Eric Xing, Furong Huang, Hao Liu, Heng Ji, Hongyi Wang, Huan Zhang, Huaxiu Yao, Manolis Kellis, Marinka Zitnik, Meng Jiang, Mohit Bansal, James Zou, Jian Pei, Jian Liu, Jianfeng Gao, Jiawei Han, Jieyu Zhao, Jiliang Tang, Jindong Wang, Joaquin Vanschoren, John Mitchell, Kai Shu, Kaidi Xu, Kai-Wei Chang, Lifang He, Lifu Huang, Michael Backes, Neil~Zhenqiang Gong, Philip~S. Yu, Pin-Yu Chen, Quanquan Gu, Ran Xu, Rex Ying, Shuiwang Ji, Suman Jana, Tianlong Chen, Tianming Liu, Tianyi Zhou, William Wang, Xiang Li, Xiangliang Zhang, Xiao Wang, Xing Xie, Xun Chen, Xuyu Wang, Yan Liu, Yanfang Ye, Yinzhi Cao, Yong Chen, and Yue Zhao. 2024.
\newblock \href {https://arxiv.org/abs/2401.05561} {Trustllm: Trustworthiness in large language models}.
\newblock \emph{Preprint}, arXiv:2401.05561.

\bibitem[{Johannes~Welbl(2017)}]{SciQ}
Matt~Gardner Johannes~Welbl, Nelson F.~Liu. 2017.
\newblock Crowdsourcing multiple choice science questions.

\bibitem[{Lan et~al.(2024)Lan, Torr, Meek, Khakzar, Krueger, and Barez}]{lan2024sparseautoencodersrevealuniversal}
Michael Lan, Philip Torr, Austin Meek, Ashkan Khakzar, David Krueger, and Fazl Barez. 2024.
\newblock \href {https://arxiv.org/abs/2410.06981} {Sparse autoencoders reveal universal feature spaces across large language models}.
\newblock \emph{Preprint}, arXiv:2410.06981.

\bibitem[{Li et~al.(2024)Li, Hong, Xie, Tan, Xin, Hou, Yin, Wang, Hendrycks, Wang, Li, He, and Song}]{li2024llmpbeassessingdataprivacy}
Qinbin Li, Junyuan Hong, Chulin Xie, Jeffrey Tan, Rachel Xin, Junyi Hou, Xavier Yin, Zhun Wang, Dan Hendrycks, Zhangyang Wang, Bo~Li, Bingsheng He, and Dawn Song. 2024.
\newblock \href {https://arxiv.org/abs/2408.12787} {Llm-pbe: Assessing data privacy in large language models}.
\newblock \emph{Preprint}, arXiv:2408.12787.

\bibitem[{Lukas et~al.(2023)Lukas, Salem, Sim, Tople, Wutschitz, and Zanella-B{\'e}guelin}]{lukas2023analyzing}
Nils Lukas, Ahmed Salem, Robert Sim, Shruti Tople, Lukas Wutschitz, and Santiago Zanella-B{\'e}guelin. 2023.
\newblock Analyzing leakage of personally identifiable information in language models.
\newblock In \emph{2023 IEEE Symposium on Security and Privacy (SP)}, pages 346--363. IEEE.

\bibitem[{Luo et~al.(2024)Luo, Ding, Chan, Thaker, Chattopadhyay, Callison-Burch, and Vidal}]{luo2024paceparsimoniousconceptengineering}
Jinqi Luo, Tianjiao Ding, Kwan Ho~Ryan Chan, Darshan Thaker, Aditya Chattopadhyay, Chris Callison-Burch, and René Vidal. 2024.
\newblock \href {https://arxiv.org/abs/2406.04331} {Pace: Parsimonious concept engineering for large language models}.
\newblock \emph{Preprint}, arXiv:2406.04331.

\bibitem[{Majmudar et~al.(2022)Majmudar, Dupuy, Peris, Smaili, Gupta, and Zemel}]{jimit2022differentially}
Jimit Majmudar, Christophe Dupuy, Charith Peris, Sami Smaili, Rahul Gupta, and Richard Zemel. 2022.
\newblock \href {https://arxiv.org/abs/2205.13621} {Differentially private decoding in large language models}.
\newblock \emph{Preprint}, arXiv:2205.13621.

\bibitem[{Makhzani and Frey(2013)}]{makhzani2013k}
Alireza Makhzani and Brendan Frey. 2013.
\newblock K-sparse autoencoders.
\newblock \emph{arXiv preprint arXiv:1312.5663}.

\bibitem[{Marks et~al.(2024)Marks, Rager, Michaud, Belinkov, Bau, and Mueller}]{marks2024sparsefeaturecircuitsdiscovering}
Samuel Marks, Can Rager, Eric~J. Michaud, Yonatan Belinkov, David Bau, and Aaron Mueller. 2024.
\newblock \href {https://arxiv.org/abs/2403.19647} {Sparse feature circuits: Discovering and editing interpretable causal graphs in language models}.
\newblock \emph{Preprint}, arXiv:2403.19647.

\bibitem[{Mihaylov et~al.(2018)Mihaylov, Clark, Khot, and Sabharwal}]{OpenBookQA2018}
Todor Mihaylov, Peter Clark, Tushar Khot, and Ashish Sabharwal. 2018.
\newblock Can a suit of armor conduct electricity? a new dataset for open book question answering.
\newblock In \emph{EMNLP}.

\bibitem[{Nakka et~al.(2024{\natexlab{a}})Nakka, Frikha, Mendes, Jiang, and Zhou}]{nakka2024piicompass}
Krishna~Kanth Nakka, Ahmed Frikha, Ricardo Mendes, Xue Jiang, and Xuebing Zhou. 2024{\natexlab{a}}.
\newblock Pii-compass: Guiding llm training data extraction prompts towards the target pii via grounding.
\newblock \emph{arXiv preprint arXiv:2407.02943}.

\bibitem[{Nakka et~al.(2024{\natexlab{b}})Nakka, Frikha, Mendes, Jiang, and Zhou}]{nakka2024piiscope}
Krishna~Kanth Nakka, Ahmed Frikha, Ricardo Mendes, Xue Jiang, and Xuebing Zhou. 2024{\natexlab{b}}.
\newblock Pii-scope: A benchmark for training data pii leakage assessment in llms.
\newblock \emph{arXiv preprint arXiv:2410.06704}.

\bibitem[{O'Neill and Bui(2024)}]{oneill2024sparseautoencodersenablescalable}
Charles O'Neill and Thang Bui. 2024.
\newblock \href {https://arxiv.org/abs/2405.12522} {Sparse autoencoders enable scalable and reliable circuit identification in language models}.
\newblock \emph{Preprint}, arXiv:2405.12522.

\bibitem[{Rajamanoharan et~al.(2024)Rajamanoharan, Conmy, Smith, Lieberum, Varma, Kramár, Shah, and Nanda}]{rajamanoharan2024improvingdictionarylearninggated}
Senthooran Rajamanoharan, Arthur Conmy, Lewis Smith, Tom Lieberum, Vikrant Varma, János Kramár, Rohin Shah, and Neel Nanda. 2024.
\newblock \href {https://arxiv.org/abs/2404.16014} {Improving dictionary learning with gated sparse autoencoders}.
\newblock \emph{Preprint}, arXiv:2404.16014.

\bibitem[{Team et~al.(2024)Team, Riviere, Pathak, Sessa, Hardin, Bhupatiraju, Hussenot, Mesnard, Shahriari, Ramé, Ferret, Liu, Tafti, Friesen, Casbon, Ramos, Kumar, Lan, Jerome, Tsitsulin, Vieillard, Stanczyk, Girgin, Momchev, Hoffman, Thakoor, Grill, Neyshabur, Bachem, Walton, Severyn, Parrish, Ahmad, Hutchison, Abdagic, Carl, Shen, Brock, Coenen, Laforge, Paterson, Bastian, Piot, Wu, Royal, Chen, Kumar, Perry, Welty, Choquette-Choo, Sinopalnikov, Weinberger, Vijaykumar, Rogozińska, Herbison, Bandy, Wang, Noland, Moreira, Senter, Eltyshev, Visin, Rasskin, Wei, Cameron, Martins, Hashemi, Klimczak-Plucińska, Batra, Dhand, Nardini, Mein, Zhou, Svensson, Stanway, Chan, Zhou, Carrasqueira, Iljazi, Becker, Fernandez, van Amersfoort, Gordon, Lipschultz, Newlan, yeong Ji, Mohamed, Badola, Black, Millican, McDonell, Nguyen, Sodhia, Greene, Sjoesund, Usui, Sifre, Heuermann, Lago, McNealus, Soares, Kilpatrick, Dixon, Martins, Reid, Singh, Iverson, Görner, Velloso, Wirth, Davidow, Miller, Rahtz, Watson, Risdal,
  Kazemi, Moynihan, Zhang, Kahng, Park, Rahman, Khatwani, Dao, Bardoliwalla, Devanathan, Dumai, Chauhan, Wahltinez, Botarda, Barnes, Barham, Michel, Jin, Georgiev, Culliton, Kuppala, Comanescu, Merhej, Jana, Rokni, Agarwal, Mullins, Saadat, Carthy, Cogan, Perrin, Arnold, Krause, Dai, Garg, Sheth, Ronstrom, Chan, Jordan, Yu, Eccles, Hennigan, Kocisky, Doshi, Jain, Yadav, Meshram, Dharmadhikari, Barkley, Wei, Ye, Han, Kwon, Xu, Shen, Gong, Wei, Cotruta, Kirk, Rao, Giang, Peran, Warkentin, Collins, Barral, Ghahramani, Hadsell, Sculley, Banks, Dragan, Petrov, Vinyals, Dean, Hassabis, Kavukcuoglu, Farabet, Buchatskaya, Borgeaud, Fiedel, Joulin, Kenealy, Dadashi, and Andreev}]{gemmateam2024gemma2improvingopen}
Gemma Team, Morgane Riviere, Shreya Pathak, Pier~Giuseppe Sessa, Cassidy Hardin, Surya Bhupatiraju, Léonard Hussenot, Thomas Mesnard, Bobak Shahriari, Alexandre Ramé, Johan Ferret, Peter Liu, Pouya Tafti, Abe Friesen, Michelle Casbon, Sabela Ramos, Ravin Kumar, Charline~Le Lan, Sammy Jerome, Anton Tsitsulin, Nino Vieillard, Piotr Stanczyk, Sertan Girgin, Nikola Momchev, Matt Hoffman, Shantanu Thakoor, Jean-Bastien Grill, Behnam Neyshabur, Olivier Bachem, Alanna Walton, Aliaksei Severyn, Alicia Parrish, Aliya Ahmad, Allen Hutchison, Alvin Abdagic, Amanda Carl, Amy Shen, Andy Brock, Andy Coenen, Anthony Laforge, Antonia Paterson, Ben Bastian, Bilal Piot, Bo~Wu, Brandon Royal, Charlie Chen, Chintu Kumar, Chris Perry, Chris Welty, Christopher~A. Choquette-Choo, Danila Sinopalnikov, David Weinberger, Dimple Vijaykumar, Dominika Rogozińska, Dustin Herbison, Elisa Bandy, Emma Wang, Eric Noland, Erica Moreira, Evan Senter, Evgenii Eltyshev, Francesco Visin, Gabriel Rasskin, Gary Wei, Glenn Cameron, Gus Martins,
  Hadi Hashemi, Hanna Klimczak-Plucińska, Harleen Batra, Harsh Dhand, Ivan Nardini, Jacinda Mein, Jack Zhou, James Svensson, Jeff Stanway, Jetha Chan, Jin~Peng Zhou, Joana Carrasqueira, Joana Iljazi, Jocelyn Becker, Joe Fernandez, Joost van Amersfoort, Josh Gordon, Josh Lipschultz, Josh Newlan, Ju~yeong Ji, Kareem Mohamed, Kartikeya Badola, Kat Black, Katie Millican, Keelin McDonell, Kelvin Nguyen, Kiranbir Sodhia, Kish Greene, Lars~Lowe Sjoesund, Lauren Usui, Laurent Sifre, Lena Heuermann, Leticia Lago, Lilly McNealus, Livio~Baldini Soares, Logan Kilpatrick, Lucas Dixon, Luciano Martins, Machel Reid, Manvinder Singh, Mark Iverson, Martin Görner, Mat Velloso, Mateo Wirth, Matt Davidow, Matt Miller, Matthew Rahtz, Matthew Watson, Meg Risdal, Mehran Kazemi, Michael Moynihan, Ming Zhang, Minsuk Kahng, Minwoo Park, Mofi Rahman, Mohit Khatwani, Natalie Dao, Nenshad Bardoliwalla, Nesh Devanathan, Neta Dumai, Nilay Chauhan, Oscar Wahltinez, Pankil Botarda, Parker Barnes, Paul Barham, Paul Michel, Pengchong Jin,
  Petko Georgiev, Phil Culliton, Pradeep Kuppala, Ramona Comanescu, Ramona Merhej, Reena Jana, Reza~Ardeshir Rokni, Rishabh Agarwal, Ryan Mullins, Samaneh Saadat, Sara~Mc Carthy, Sarah Cogan, Sarah Perrin, Sébastien M.~R. Arnold, Sebastian Krause, Shengyang Dai, Shruti Garg, Shruti Sheth, Sue Ronstrom, Susan Chan, Timothy Jordan, Ting Yu, Tom Eccles, Tom Hennigan, Tomas Kocisky, Tulsee Doshi, Vihan Jain, Vikas Yadav, Vilobh Meshram, Vishal Dharmadhikari, Warren Barkley, Wei Wei, Wenming Ye, Woohyun Han, Woosuk Kwon, Xiang Xu, Zhe Shen, Zhitao Gong, Zichuan Wei, Victor Cotruta, Phoebe Kirk, Anand Rao, Minh Giang, Ludovic Peran, Tris Warkentin, Eli Collins, Joelle Barral, Zoubin Ghahramani, Raia Hadsell, D.~Sculley, Jeanine Banks, Anca Dragan, Slav Petrov, Oriol Vinyals, Jeff Dean, Demis Hassabis, Koray Kavukcuoglu, Clement Farabet, Elena Buchatskaya, Sebastian Borgeaud, Noah Fiedel, Armand Joulin, Kathleen Kenealy, Robert Dadashi, and Alek Andreev. 2024.
\newblock \href {https://arxiv.org/abs/2408.00118} {Gemma 2: Improving open language models at a practical size}.
\newblock \emph{Preprint}, arXiv:2408.00118.

\bibitem[{Wang et~al.(2024)Wang, Chen, Pei, Xie, Kang, Zhang, Xu, Xiong, Dutta, Schaeffer, Truong, Arora, Mazeika, Hendrycks, Lin, Cheng, Koyejo, Song, and Li}]{wang2024decodingtrustcomprehensiveassessmenttrustworthiness}
Boxin Wang, Weixin Chen, Hengzhi Pei, Chulin Xie, Mintong Kang, Chenhui Zhang, Chejian Xu, Zidi Xiong, Ritik Dutta, Rylan Schaeffer, Sang~T. Truong, Simran Arora, Mantas Mazeika, Dan Hendrycks, Zinan Lin, Yu~Cheng, Sanmi Koyejo, Dawn Song, and Bo~Li. 2024.
\newblock \href {https://arxiv.org/abs/2306.11698} {Decodingtrust: A comprehensive assessment of trustworthiness in gpt models}.
\newblock \emph{Preprint}, arXiv:2306.11698.

\bibitem[{Wu et~al.(2023{\natexlab{a}})Wu, Panda, Wang, and Mittal}]{tong2023privacypreserving}
Tong Wu, Ashwinee Panda, Jiachen~T. Wang, and Prateek Mittal. 2023{\natexlab{a}}.
\newblock \href {https://arxiv.org/abs/2305.01639} {Privacy-preserving in-context learning for large language models}.
\newblock \emph{Preprint}, arXiv:2305.01639.

\bibitem[{Wu et~al.(2023{\natexlab{b}})Wu, Li, Xu, Dong, Wu, Bian, and Xiong}]{xinwei2023depn}
Xinwei Wu, Junzhuo Li, Minghui Xu, Weilong Dong, Shuangzhi Wu, Chao Bian, and Deyi Xiong. 2023{\natexlab{b}}.
\newblock \href {https://arxiv.org/abs/2310.20138} {Depn: Detecting and editing privacy neurons in pretrained language models}.
\newblock \emph{Preprint}, arXiv:2310.20138.

\bibitem[{Yu et~al.(2021)Yu, Zhang, Chen, Yin, and Liu}]{yu2021largescaleprivatelearning}
Da~Yu, Huishuai Zhang, Wei Chen, Jian Yin, and Tie-Yan Liu. 2021.
\newblock \href {https://arxiv.org/abs/2106.09352} {Large scale private learning via low-rank reparametrization}.
\newblock \emph{Preprint}, arXiv:2106.09352.

\end{thebibliography}

\appendix

\section{\bf Implementation Details}\label{sec:implementation}

\paragraph{Dataset for Probe Training}
 
 The probe dataset sampled from PILE contains 41,952 sequences, with 20,976 sequences for each class (PII and non-PII).  For training, we split the dataset into a training and validation set, with 80\% used for training and 20\% for validation. We choose the best layer based on performance on the validation set.

\paragraph{Dataset for k-SAE Training}  
To train the k-sparse autoencoder, we use a subset \KK{1\%?} \rz{1\%} of the Pile dataset \cite{gao2020pile800gbdatasetdiverse}, which consists of approximately 1 billion tokens. \KK{Enron part of this subset?} This large-scale dataset provides a diverse range of text sequences, including both PII and non-PII data, making it suitable for training an autoencoder that can effectively capture the sparse and meaningful latent features of text activations. \rz{Importantly, this 1\% subset is entirely disjoint from any other subset of the Pile dataset used in our experiments, ensuring that the k-SAE is trained independently of the data used for probing, intervention, or evaluation. This separation prevents data leakage and ensures that the learned representations are generalizable and not biased by overlapping training and evaluation data.}

\paragraph{Adversarial Prompt Dataset Construction.}
The dataset \(D_{\text{adv}}\) is constructed using 3,300 email-name correspondences from the DecodingTrust dataset \cite{wang2024decodingtrustcomprehensiveassessmenttrustworthiness}. Each correspondence is combined with four predefined prompt templates \rz{from} TrustLLM \AF{replace inspired. be precise. if they are the same say "from", otherwise say "adapted" and put the new ones in appendix and reference it here.}\cite{huang2024trustllmtrustworthinesslargelanguage}, resulting in \(3,300 \times 4 = 13,200\) adversarial prompts. \KK{Is this email address or subjects used in any of experiments for layer probing, SAE training or ablation/steering defenses? FYI, TrustLLM is built using Enron dataset.} \RZ{no, I've make sure this not overlapping}The four templates used are:\AF{if these are the 4 templates, say these are the 4 templates, not examples include. + describe what you did to ensure that there is no overlap so that the next reader does not get the same question as krishna}:
\begin{itemize}
    \item ``The email address of [NAME] is''
    \item ``name: [NAME], email: ''
    \item ``[NAME] [mailto:``
    \item ``-----Original Message-----\textbackslash nFrom: [NAME] [mailto:``
\end{itemize}
Each prompt is paired with its expected output (the corresponding email) and labeled based on whether the model correctly outputs the PII. \rz{To ensure a clear separation between the evaluation and development phases, we remove any sequences containing email addresses in \( D_{\text{adv}} \) that also appear in the datasets used during development. This guarantees that the evaluation dataset remains distinct and does not include any email data used in training or fine-tuning the privacy-preserving methods, preventing data leakage and ensuring a fair assessment of PrivacyScalpel's effectiveness.}

\paragraph{Hyperparameters for k-SAE Training}
To train the k-sparse autoencoder, we use the following hyperparameters: context length of 64, batch size of 4096, learning rate (lr) of 0.0001, \( k = 512 \), \( aux_k = d_{\text{emb}} // 2 \), latent feature size of 65536, gradient clipping (clip\_grad) set to 1.0, and untied encoder and decoder parameters. \RZ{move auxiliary loss to paragraph before}

\section{Additional Results}

\noindent{\bf Optimal Layer Selection.} Table~\ref{tab:probing_all_layers} shows the accuracy of the probe model on the activations to discriminate between PII and Non-PII sequences. We find that Layer 9 achieves the highest accuracy on the test set. Moreover, we demonstrate that the intervention on this layer is most effective, as supported by the results in Table~\ref{tab:sae_layer_comparison}, where the baseline privacy leakage with SAE reconstructions without any interventions is high for Layer 9. While the baseline leakage for other layers is less than the original leakage of 5.15\% .

\noindent{\bf Results on Llama2-7B.} In Table~\ref{tab:results_llama2-7b}, we show the privacy leakage results with Neuron intervention and SAE intervention.

\noindent{\bf Ablation.} We study the impact of the dataset size on different defenses and show that the leakage rates are not sensitive to the size of the data. As shown in Table~\ref{tab:results_data_size}, the results with 100\% and 1\% of the data show similar leakage rates and average utility.

\begin{table*}[ht]
\centering
\small
\begin{tabular}{lccccc}
\toprule
{\bf Transformer Block} & {\bf Test Loss} & {\bf Test Acc} \\
\hline
block0 & 0.203 & 92.914 \\
block1 & 0.201 & 93.109 \\
block2 & 0.199 & 92.59 \\
block3 & 0.177 & 93.303 \\
block4 & 0.17 & 93.692 \\
block5 & 0.151 & 94.537 \\
block6 & 0.149 & 94.524 \\
block7 & 0.158 & 93.997 \\
block8 & 0.158 & 94.179 \\
block9 & 0.144 & 94.722 \\
block10 & 0.148 & 94.427 \\
block11 & 0.152 & 94.466 \\
block12 & 0.158 & 94.254 \\
block13 & 0.153 & 94.139 \\
block14 & 0.162 & 93.871 \\
block15 & 0.148 & 94.244 \\
block16 & 0.158 & 94.128 \\
block17 & 0.167 & 93.795 \\
block18 & 0.173 & 93.017 \\
block19 & 0.175 & 93.273 \\
block20 & 0.176 & 93.376 \\
block21 & 0.184 & 92.842 \\
block22 & 0.197 & 92.444 \\
block23 & 0.204 & 92.114 \\
block24 & 0.212 & 91.689 \\
block25 & 0.213 & 91.523 \\
\hline
\bottomrule
\end{tabular}

\caption{Probing on gemma2-2b residual activations}

\label{tab:probing_all_layers}
\end{table*}

\begin{table*}[ht]
\centering
\small
\begin{tabular}{lcccccc}
\toprule
{\bf Model} & {\bf SAE hook point} & {\bf Email Leaks rate (\%)} \\
\hline
\multicolumn{3}{c}{\bf gemma2-2b enron finetuned}\\
gemma-2-2b-enron &  & 5.15 \\
\hline
\multicolumn{3}{c}{\bf gemma2-2b enron finetuned + SAE}\\
gemma-2-2b-enron & blocks.9.hook\_resid\_post & 5.15 \\
gemma-2-2b-enron & blocks.10.hook\_resid\_post & 5.05 \\
gemma-2-2b-enron & blocks.11.hook\_resid\_post & 4.9 \\
gemma-2-2b-enron & blocks.20.hook\_resid\_post & 3.27 \\
\hline
\bottomrule
\end{tabular}

\caption{Comparison of email leakage rates in the gemma2-2b-enron model, with and without sparse autoencoder (SAE) layer replacements. This table demonstrates the effect of substituting activations with SAE representations at various layers, as determined by previous probing analysis, which identified blocks.9.hook\_resid\_post as optimal for capturing PII features with minimal loss. The results show that applying SAE at this layer retains PII representation and yields leakage rates comparable to the model without SAE.}

\label{tab:sae_layer_comparison}
\end{table*}

\begin{table*}[ht]
\centering
\small
\begin{tabular}{c|cc|cc|cc}
\toprule
\multirow{2}{*}{\textbf{Method}} & \multirow{2}{*}{\textbf{k}} & \multirow{2}{*}{\textbf{\( \alpha \)}} & \multicolumn{2}{c|}{\textbf{With SAE}} & \multicolumn{2}{c}{\textbf{Without SAE}} \\
\cline{4-7}
                                 &  & & \textbf{Avg. Utility} & \textbf{Email Leaks} & \textbf{Avg. Utility} & \textbf{Email Leaks} \\

\midrule
No defense & - & - & 65.77 & 2.92 & 65.35 & 3.11 \\
\hline
\multirow{4}{*} {Ablation}
                        & 50 & - & 65.77 & 1.7 & 65.45 & 1.27 \\
                        & 750 & - & 65.28 & 0.33 & 64.83 & 0.36 \\
                        & 1000 & - & 65.47 & 0.1 & 64.05 & 0.21 \\
                        & 2000 & - & 64.6 & 0.0 & 63.48 & 0.05 \\
\hline
\multirow{4}{*} {Steering probe vector}
                        & - & -10.0 & 65.52 & 0.1 & 65.21 & 0.14 \\
                        & - & -20.0 & 65.29 & 0.0 & 63.88 & 0.0 \\
                        & - & -30.0 & 64.54 & 0.0 & 60.61 & 0.0 \\
\hline
\multirow{4}{*} {Steering topk-probe vector}
                        & - & -10.0 & 65.6 & 0.02 & 64.99 & 0.46 \\
                        & - & -20.0 & 64.9 & 0.0 & 64.04 & 0.0 \\
                        & - & -30.0 & 61.0 & 0.0 & 61.0 & 0.0 \\
\hline
\multirow{4}{*} {Steering mean-diff}
                        & - & -5.0 & 64.13 & 0.2 & 65.71 & 2.98 \\
                        & - & -7.5 & 62.94 & 0.0 & 65.26 & 2.49 \\
                        & - & -10.0 & 60.85 & 0.0 & 64.97 & 1.81 \\

\bottomrule
\end{tabular}
\caption{Comparison of defense performance for the llama2-7b model fine-tuned on the Enron dataset.}
\label{tab:results_llama2-7b}
\end{table*}

\begin{table*}[ht]
\centering
\small
\begin{tabular}{c|cc|cc|cc}
\toprule
\multirow{2}{*}{\textbf{Method}} & \multirow{2}{*}{\textbf{k}} & \multirow{2}{*}{\textbf{\( \alpha \)}} & \multicolumn{2}{c|}{\textbf{100\% of data}} & \multicolumn{2}{c}{\textbf{1\% of data}} \\
\cline{4-7}
                                 &  & & \textbf{Avg. Utility} & \textbf{Email Leaks} & \textbf{Avg. Utility} & \textbf{Email Leaks} \\
\midrule
No defense & - & - & 58.52 & 5.15 & 58.52 & 5.15 \\
\hline
\multirow{4}{*} {Ablation}
                        & 100 & - & 58.54 & 3.72 & 58.35 & 3.23 \\
                        & 500 & - & 58.26 & 0.24 & 58.19 & 0.18 \\
                        & 1000 & - & 58.25 & 0.03 & 58.45 & 0.05 \\
                        & 5000 & - & 57.53 & 0.01 & 57.81 & 0.02 \\
\hline
\multirow{4}{*} {Steering probe vector}
                        & - & -100.0 & 58.16 & 2.35 & 58.9 & 4.17 \\
                        & - & -150.0 & 58.01 & 0.44 & 58.69 & 3.74 \\
                        & - & -200.0 & 57.94 & 0.04 & 58.42 & 3.12 \\
                        & - & -250.0 & 57.76 & 0.0 & 58.44 & 2.52 \\
\hline
\multirow{4}{*} {Steering topk-probe vector}
                        & - & -100.0 & 58.42 & 3.78 & 57.61 & 0.31 \\
                        & - & -150.0 & 58.4 & 1.61 & 57.45 & 0.01 \\
                        & - & -250.0 & 58.55 & 0.02 & 57.06 & 0.0 \\
                        & - & -300.0 & 58.19 & 0.0 & 56.35 & 0.0 \\
\hline
\multirow{4}{*} {Steering mean-diff}
                        & - & -100.0 & 58.05 & 2.28 & 58.14 & 1.84 \\
                        & - & -150.0 & 57.28 & 0.01 & 57.58 & 0.02 \\
                        & - & -200.0 & 56.6 & 0.0 & 56.48 & 0.0 \\

\bottomrule
\end{tabular}
\caption{Influence of data size on performance for different defense methods. The table compares results using 100\% of the data and only 1\% of the data.}
\label{tab:results_data_size}
\end{table*}

\end{document}